\documentclass{article}

\usepackage[dblblindworkshop, final]{neurips_2025}

\workshoptitle{Data on the Brain \& Mind}

\usepackage[utf8]{inputenc} 
\usepackage[T1]{fontenc}    
\usepackage{hyperref}       
\usepackage{url}            
\usepackage{booktabs}       
\usepackage{amsfonts}       
\usepackage{graphicx}       
\usepackage{nicefrac}       
\usepackage{microtype}      
\usepackage{xcolor}         
 \usepackage{amsmath}
 \usepackage{float}

\title{LearnAD: Learning Interpretable Rules for Brain Networks in
Alzheimer’s Disease Classification}

\author{%
    Thomas Andrews \\
  Department of Computing\\
  Imperial College London\\
  \texttt{ta4218@imperial.ac.uk} \\
  \And
  Alessandra Russo \\
Department of Computing\\
  Imperial College London\\
  \texttt{a.russo@imperial.ac.uk} \\
  \AND
  Sara Ahmadi-Abhari \\
    School of Public Health\\
  Imperial College London\\
  \texttt{s.ahmadi-abhari@imperial.ac.uk} \\
  \And
  Mark Law \\
  ILASP LTD \\
\texttt{mark@ilasp.com} \\
}

\begin{document}

\maketitle

\begin{abstract}
We introduce LearnAD, a neuro-symbolic method for predicting Alzheimer’s disease from brain magnetic resonance imaging data, learning fully interpretable rules. LearnAD applies statistical models, Decision Trees, Random Forests, or GNNs to identify relevant brain connections, and then employs FastLAS to learn global rules. Our best instance outperforms Decision Trees, matches Support Vector Machine accuracy, and performs only slightly below Random Forests and GNNs trained on all features, all while remaining fully interpretable. Ablation studies show that our neuro-symbolic approach improves interpretability with comparable performance to pure statistical models. LearnAD demonstrates how symbolic learning can deepen our understanding of GNN behaviour in clinical neuroscience.
\end{abstract}

\section{Introduction}
\label{sec:intro}
Alzheimer's Disease (AD) is a progressive neurodegenerative disease characterised by the accumulation of extracellular amyloid-$\beta$ plaques with subsequent intracellular tau neurofibrillary tangles \citep{Braak1991}. However, these misfolded proteins appear in the brain prior to presentation of clinical and cognitive symptoms, such as cognitive decline and dementia \citep{Ossenkoppele2022,Yang2021}. Therefore, an improved understanding of the underlying mechanisms of the pathology's progression and spread is essential for the early diagnosis of AD and for identifying therapeutic targets. One avenue is to understand associations between structural brain networks and disease progression and manifestation. In this work, we utilise the Alzheimer's Disease Neuroimaging Study (ADNI)~\citep{JackJr2008} (\url{https://adni.loni.usc.edu/}) to learn rules that distinguish between individuals who are cognitively normal (CN) and those who have AD.

Statistical machine learning methods have been used to predict Alzheimer’s disease (AD) from structural and functional brain magnetic resonance imaging (MRI) \citep{Gam_Disentangled_MICCAI2024,Wan_Selfguided_MICCAI2024}. Early work relied on handcrafted features with classifiers such as Support Vector Machines (SVMs), Decision Trees (DTs), and Random Forests (RFs) \citep{Gerardin2009,dt_ad,Sarica2017}. DTs are generally more interpretable than SVMs and RFs, making them preferable when explanations of learned knowledge are required \citep{BarredoArrieta2020}. RFs aggregate predictions across multiple DTs, improving generalisation and typically outperforming DTs, albeit with reduced transparency \citep{BarredoArrieta2020}.

Recently, with the availability of large datasets such as ADNI, deep convolutional neural networks (CNNs) have been shown to learn discriminative patterns from minimally preprocessed MRI data \citep{ZhouYanjieandLi2023}. Despite strong performance, particularly in cross-sectional AD, these models remain black boxes. Their prediction processes are non-transparent, reducing clinician trust and hindering verification that they learn true disease markers rather than confounds. Graph neural networks (GNNs) have been increasingly studied in the healthcare domain \citep{SchlichtkrullMichaelandKipf2018,10.5555/3305381.3305512}. They have emerged as a promising paradigm for modelling AD, offering a natural way to capture the brain’s network organisation. Typically, nodes represent regions of interest (ROIs), while edges encode structural or functional connectivity derived from MRIs. Message-passing enables the use of local neighbourhoods for information flow, thus making predictions heavily influenced by the graph's natural local structure when learning global representations \citep{Tang2023}. In a clinical context, GNNs learn structural patterns that are predictive of AD. However, when interpreting learned models, it is unclear what contribution specific node features, edges, and subgraphs make in the inference process~\citep{Dehmamy2019}. 

Addressing interpretability when learning predictive models for AD remains an open challenge.  Symbolic machine learning offers an alternative learning paradigm with formal guarantees of learning from structured data and domain knowledge, with sound and verifiable decision-making \cite{Law23, LawRBB020}. Learned predictive models, represented in logical form, are inherently transparent and human-interpretable. However, the unstructured nature of MRI data and the complexity of predicting AD challenge their applicability and scalability, requiring the need for new methods that can enable symbolic learning from unstructured MRI data in a controlled search space, without reducing their predictive performance. Automatic engineering of relevant structured features from unstructured MRI data can aid in this endeavour.



In this work, we propose LearnAD, a new neuro-symbolic approach that combines statistical machine learning with symbolic learning to learn global interpretable rules for predicting AD from structural MRI of the brain. Firstly, a statistical machine learning component is trained on labelled features extracted from structural MRI data. The features most relevant in the classification of AD patients are selected from this trained model. In the second step, we use a symbolic machine learning system, called FastLAS~\citep{Law2020}, known for learning interpretable rule-based models from noisy (unstructured) data~\cite{Cunnington2023}. Symbolic examples are automatically generated from the labelled patient data (AD versus CN) using the selected patient instance values of the relevant features selected in the first step, but represented as structured contextual information. The selected features are also used to constrain the search space for the symbolic learner. The learned predictive model, generated by FastLAS, generalises across the different local instance-level connections of the brain and determines the most relevant affected brain connections responsible for differentiating AD from CN, and a semi-parametric bound over these regions. 

We consider three different instances of our approach, using respectively different statistical machine learning models, DT, RF, and GNN, for learning the most relevant local subgraphs of brain regions. This is in order to evaluate which of these statistical machine learning models is most effective for feature selection when combined with the symbolic learner. We use cross-validation to evaluate the three different instances and show that the accuracy of the predictive model that uses features extracted from the DT outperforms the other two instances. We use a DT, RF, and GNN trained on the full set of features as baseline models to evaluate if the generalisation power of the symbolic learner would achieve similar accuracy with limited features. 

Our results show that our best performing instance outperforms the DT, is as accurate as an SVM, and slightly underperforms against the RF and GNN when trained over the full set of features. In all cases, our approach is fully interpretable. Finally, we perform an ablation study to evaluate the benefit of a combined neuro-symbolic approach versus its respective pure statistical machine learning counterpart. In this case, we consider only the DT and RF counterpart models, as the GNN would not be trainable on the sparse networks. Our ablation study shows that our combined neuro-symbolic approach outperforms the DT, and slightly underperforms with respect to an RF trained on these same selected features, but the interpretability of our symbolically learned predictive model is much higher than the RF-trained model. 

%


The paper is structured as follows. In Section~\ref{sec:background}, we introduce the techniques that our approach builds upon. Section~\ref{sec:data} offers a description of the preprocessing of our ADNI dataset that we use in our experiments, followed by Section~\ref{sec:methods}, which presents our neuro-symbolic approach. The results of our experiments are discussed in Section~\ref{sec:discussion}. Section~\ref{sec:related} summarises related work in the area of AD prediction, and finally Section~\ref{sec:conclusions} concludes the paper, suggesting future research directions.

\section{Background}
\label{sec:background}
In this section, we cover two main topics necessary to understand our proposed approach. We briefly describe the statistical machine learning methods that we use to instantiate our approach, and we present the Learning from Answer Sets framework~\cite{LawRB18} and its current state-of-the-art (SOTA) system, FastLAS~\cite{LawRBB020}.

\subsection{Statistical Machine Learning}
In this paper, we focus on three machine learning approaches, DT, RF, and GNN, and their use in binary classification tasks. Classification and Regression Tree (CART) is among the most common methods for learning binary classification decision trees \citep{Breiman1984}. Given a labelled dataset $D=\{(X,y)\}$, the CART algorithms learn a binary tree model that covers the labelled data. Internal nodes are decision points on features in $X$. Each branch of the learned tree is a decision rule, learned from the data features to maximise the coverage of the given labelled data. 

CART is a recursive binary algorithm; at each node it chooses a feature and a split point that best separates the data by reducing the Gini Impurity: the measure of how mixed (or how homogeneous) the classes are at that node. Specifically, it measures the probability that a randomly chosen sample from that node would be incorrectly classified if it is labelled according to the class distribution in that node.
Formally, the Gini Impurity is defined as follows:
\begin{equation}\label{eq:gini}
I_G = 1 - \sum_{j=1}^{C} {p_j}^2.
\end{equation}
where $C$ is the number of classes, and $p_j$ is the proportion of the samples at the node belonging to class $j$. The algorithm aims to partition the feature space into regions that are as ``pure'' as possible (i.e., have a minimal value of Gini Impurity), so that each region ideally contains samples of only one class. 
%
%
RF are extensions of DT, employing a number of DTs, trained on a bootstrapped subsample of features in $X$, utilising averaging to improve the predictive accuracy of the model \citep{Breiman2001}.

Graph Neural Networks (GNNs) are a class of deep learning models designed to work directly on graph-structured data, where information is represented as nodes (entities) and edges (relationships). One of the most widely adopted GNNs is Graph Convolutional Networks (GCN)~\cite{kipf2017semisupervisedclassificationgraphconvolutional}.
The objective of GCNs is to learn node embeddings and graph representations to perform downstream classification. 
A GCN is a binary undirected graph of $m$ nodes with $n$ features per node. It is represented as an adjacency matrix, $A \in \{0,1\}^{m \times m}$, with a feature matrix $X \in \mathbb{R}^{m \times n}$. For a weighted graph, with real-numbered edge weights, the adjacency matrix is extended to $A \in \mathbb{R}^{m \times m}$. 
During training, a single GCN layer updates node embeddings as 
\begin{equation}
H^{(l+1)} = \sigma(\tilde{D}^{-1/2}\hat{A}\tilde{D}^{-1/2}H^{(l)}W^{(l)})
\end{equation} 
where $\hat{A}=A+I$, $\tilde{D}^{-1/2}$ is the diagonal degree matrix, $W^{(l)}$ is a learnable weight matrix, and $\sigma$ is the activation. This transformation aggregates information at each node from its local neighbourhood, producing updated node embeddings layer by layer. After a readout step that pools node embeddings into a global graph representation, a multi-layer perceptron produces the final classification.
%

To extract edges relevant to the prediction, we implement \textit{GNNExplainer} \citep{ying2019gnnexplainergeneratingexplanationsgraph}. \textit{GNNExplainer} is chosen due to its simplicity, speed, and its model-agnostic design applicable to any GNN architecture. Given a target classification, in a graph, $G$, \textit{GNNExplainer} aims to find a subgraph $G_s \in G$, and a reduced subset of features $X_s = \{x_i | v_i \in G_s\}$ that are important in a prediction, $y$. Importance is defined with respect to mutual information (MI), such that the \textit{GNNExplainer} aims to maximise the function:

\begin{equation}\label{eq:maxgs}
\max_{G_s} (Y,(G_s,X_s)) = H(Y) - H(Y | G = G_s, X = X_s).
\end{equation}

We refer the reader to the original paper for optimisation details. For now, we summarise the final result; a soft mask is learned over the adjacency matrix for subgraphs $G_s$, $A_s  \in [0,1]^{n \times n}$. 

\subsection{Learning from Answer Sets}
Learning from Answer Sets (LAS) \cite{Law2020} is a symbolic machine learning \cite{Muggleton1991} paradigm for learning Answer Set Programs (ASP). First, we briefly introduce ASP programs. 

Answer Set Programming is a symbolic formalism for representing knowledge and performing inference. Within the scope of this paper, ASP programs are constituted of \emph{normal} rules and \emph{constraints}. The former are of the form $h$ :- $b_1$,...,$b_m$, not $c_1$,...,not $c_m$, where $h$ is the \emph{head} and $b_1$,...,$b_m$, not $c_1$,...,not $c_m$ is collectively the \emph{body} of the rule, not represents negation as failure. A normal rule reads as ``$h$ is true if all $b_i$ are true and none of the $c_j$ are proved to be true''. Constraints take the form $:\!-\; b_1,\cdots,b_n$ and have the effect of ruling out solutions where the body is satisfied. The semantics of ASP programs is based on the notion of Herbrand interpretations. The Herbrand interpretation of a given program $P$ is a set of ground atoms constructed from the relations and objects that appear in $P$.  Given an ASP program $P$ and a Herbrand interpretation $I$, the reduct program $P^I$ can be constructed using the following three steps: 1. Eliminate all rules containing negated atoms that appear in $I$; 2. Strip all negated atoms from the body of the remaining rules; 3. Replace constraint heads with $\perp$ (meaning falsity). The interpretation $I$ is an answer set (i.e. model) of the program $P$, if it is the minimal model of the reduct program $P^I$. A partial interpretation is expressed as the pair $e_{pi}=\langle e^{inc},e^{exc}\rangle$, where $e^{inc}$ and $e^{exc}$ contain ground atoms for inclusions and exclusions respectively. An interpretation $I$ \emph{extends} a partial interpretation $e_{pi}$ when $e^{inc} \subseteq I$ and $e^{exc} \cap I = \emptyset$.

A Learning from Answer Sets task is a tuple $\langle B,S_M,E\rangle$, where $B$ is an ASP called \textit{background knowledge}, $S_M$ is a set of rules known as a \textit{hypothesis space}, and $E$ is a set of \textit{examples}. The hypothesis space is specified by means of a set of \textit{mode declarations}, defining which predicates can appear in the head or body of a rule. In this work, we use the SOTA LAS system FastLAS~\cite{LawRBB020}. Examples in FastLAS are referred to as \textit{weighted context-dependent partial interpretation} (WCDPI). They are of the form $\langle e_{id}, e_{pen}, \langle e^{inc},e^{exc}\rangle,e_{ctx}\rangle$, where $e_{id}$ is an identifier for the example, $e_{pen}$ is a penalty for not covering the example, $\langle e^{inc},e^{exc}\rangle$ is a partial interpretation and $e_{ctx}$ is an ASP known as the \textit{context} for the example used to provide example-specific information, such as example features. The solution of a given LAS task is called a \emph{hypothesis} and it is an ASP program $H \subseteq S_M$. $H$ is said to \emph{cover} an example $e$ if there exists an answer set of $B \cup e_{\text{ctx}} \cup H$ that contains every atom in $e^{inc}$ and no atom in $e^{exc}$. The score of a hypothesis $H$ is the sum of the length of $H$ (in terms of the number of atoms that appear in $H$) and the penalties of uncovered examples. An optimal solution is a hypothesis $H \subset S_M$ with minimal score.




FastLAS's scalability is rooted in the computation of an opt-sufficient-subset of the given hypothesis space that guarantees the existence of an optimal solution. It also allows for meta-bias over the search space and domain-specific scoring functions as mechanisms for controlling the size of the search space and enabling scalability.  The reader is referred to~\cite{LawRBB020} for further details on FastLAS. 

\section{Dataset and Processing}
\label{sec:data}
\subsection{ADNI}
Data were extracted as follows: we retrieved all available structural connectomes from ADNI2 and ADNI3 (phases two and three of ADNI). In keeping with case-control studies, each AD patient was matched to a CN patient by age and sex. The resulting dataset is balanced, with 152 CN patients and 152 AD patients (304 total) who were matched by age (CN: 76.7; AD: 76.4) and sex (F: 120; M: 184). Patients had images recorded at multiple time points; we extracted their latest available image. Patients labelled CN maintain a healthy brain throughout the duration of the study. Meanwhile, patients who are labelled AD have, at the time of imaging, been diagnosed with AD.

\subsection{Connectome Extraction}
Structural connectomes were extracted with Clinica, a software platform for clinical neuroscience research \citep{Routier2021}. Each patient had a T1W MRI with an associated DWI MRI. T1W MRIs were processed with the \texttt{t1-freesurfer} pipeline. This pipeline is a wrapper for different functionalities of the FreeSurfer software \citep{Fischl2012}. The processing includes segmentation of subcortical structures, extraction of cortical surfaces, cortical thickness estimation, spatial normalisation onto the FreeSurfer surface template (FsAverage), and parcellation of cortical regions~\citep{Routier2021}.

The DWI MRIs are first processed with the \texttt{dwi-preprocessing-using-t1} pipeline in Clinica. The pipeline corrects for motion, eddy currents, magnetic susceptibility, and bias field distortions \citep{Routier2021}. The pipeline utilises FSL \citep{Jenkinson2012}, ANTs \citep{Avants2014}, and MRtrix3 \citep{Tournier2019}. Finally, the connectomes are extracted with the \texttt{dwi-connectome} pipeline in Clinica. The connectomes are weighted graphs that encode the structural connections between brain regions defined with a standardised brain atlas. In this work, we use the Desikan--Killiany atlas and obtain 84 brain regions of interest (ROI), consisting of 34 cortical ROIs and 8 subcortical ROIs \citep{Desikan2006}. This pipeline utilises MRtrix3 and FreeSurfer.

\section{Methodology}
\label{sec:methods}
\subsection{Overview of the approach}
\begin{figure}[h]
\centering
\includegraphics[width=0.85\linewidth]{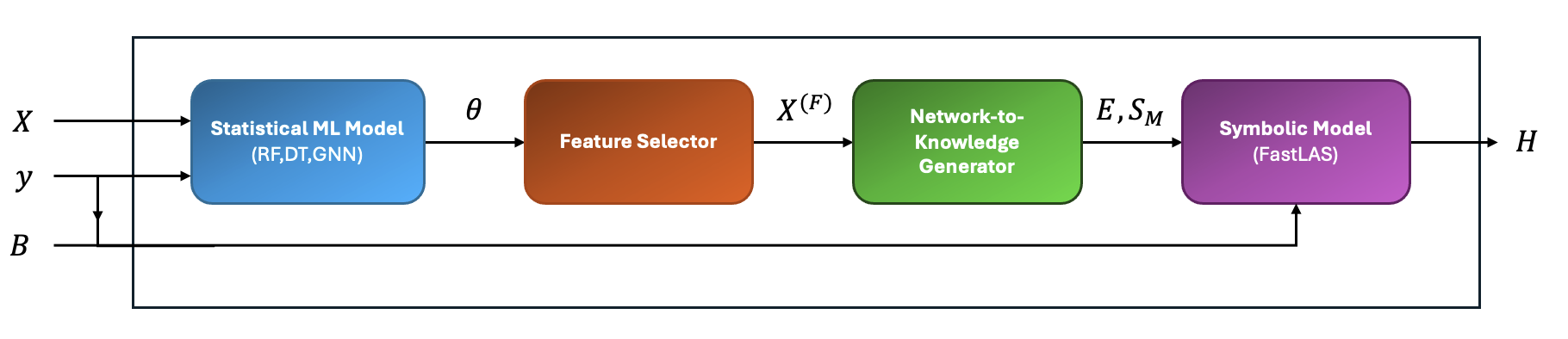}
  \caption{Architecture of our neuro-symbolic approach. The model consists of 4 components: Statistical ML Model, Feature Selector, Network-to-Knowledge Generator, and Symbolic Model. The dataset $(X,y)$ is provided to the model, along with the background knowledge, $B$. $\theta$ defines the learned parameters of the statistical models, $X^{(F)}$ are the learned relevant features. $S_M$ is the hypothesis space and $E$ are the examples. The learned hypothesis, $H$, is the output.}
  \label{fig:model}
\end{figure}

A diagram of LearnAD is shown in Figure~\ref{fig:model}. LearnAD takes as input a labelled dataset of preprocessed structural MRI data and relevant domain-specific background knowledge $B$. 

The first step of our learning approach is to train a statistical machine learning model over the dataset $\{(X,y)\}\subseteq\mathcal{X}\times\mathcal{Y}$, and select from the trained model relevant features. $\mathcal{X}$ defines the space of structural brain networks, whose representation depends on the specific statistical machine learning method used by the approach. 
$\mathcal{Y}$ describes the space of classification labels, $\mathcal{Y} =\{CN,AD\}$. The Feature Selector identifies in the trained model the most important brain connections in predicting AD. 


The selected features are subsequently input to the Network-to-Knowledge Generator (NKG) to define the learning task for the symbolic learner, i.e., construct the set of examples $E$ and the hypothesis space $S_{M}$. Finally, the FastLAS symbolic learner is used to solve the learning task and generate an optimal solution $H$ in terms of an ASP program that maximises the coverage of the set $E$ of examples. This learning step also takes into account the domain-specific background knowledge $B$.

\subsection{Statistical Classifier}
The first component of our learning task is the specification of our statistical classifier. Each data point is a symmetric weighted anatomical graph, or connectome. Modern neuroimaging techniques are unable to completely remove spurious brain connections, resulting in noisy connectomes \citep{Roberts2017}. We mask each connectome to remove these connections.

For the DT and RF, we learn  a mapping $f:\mathcal{X}\mapsto\mathcal{Y}$, that maps feature vectors $X\in\mathcal{X}$ to a label $y\in\{CN,AD\}$. Each feature vector represents the flattened lower triangular matrix of the masked connectome, with each element defining an ROI-to-ROI connection. 

The GCN is characterised by $f(X,\mathcal{E},A)$, where $X$, $\mathcal{E}$, and $A$ are the feature matrix, edge-feature matrix, and adjacency matrix. In the formulation of our GCN, $X$ is an identity matrix, a positional encoding for the ROIs. As all connectomes are masked, $A$ is the same for all data points. Although the use of edge features alone may seem counterintuitive given that all graphs share the same \( A \) and \( X \), it is precisely these values that encode the anatomical variation relevant to the task driven by connectivity patterns within each node’s local neighbourhood. We learn the mapping $f : (X,\mathcal{E},A) \mapsto \hat{y} $ where $\hat{y}\in \mathcal{Y}$ approximates $y$.

\subsection{Feature Selector}
\label{fs}
To improve the scalability of our symbolic reasoning model, we reduce the context space to $k$ ROI-to-ROI connections most relevant to the learning task. For the instances of our approach that use DT and RF as a statistical ML model, the top $k_{global}$ features (ROI-to-ROI connectivity strength) are extracted by an importance measure: the total reduction in the Gini Impurity atrributable to each feature, summed over all nodes (and averaged over all DTs for RFs). These features are ``global'' (at the model level) and therefore refer to the full dataset.

For the instance of our approach that uses a GCN classifier, the feature selector is more complex. It employs the \textit{GNNExplainer} to extract edge-level explanations for each individual training graph \( x \in \mathcal{X} \). Specifically, for each training graph, \textit{GNNExplainer} identifies a set of salient edges \( \mathcal{E}_s \subset \mathcal{E} \), where \( |\mathcal{E}_s| = k \) (controlled by limiting the explanation size to \( k_{instance} \) edges). As these explanations are at the data point level, we maintain a frequency count \( C: \mathcal{E}_s \to \mathbb{N} \) for each edge over the full set of training instances. For each $\epsilon\in\mathcal{E}$, $C(\epsilon)$ gives the number of times the edge $\epsilon$ is returned by \textit{GNNExplainer} among the top $k_{instance}$ relevant features, over the full training set. The selector returns the set $M$ (where $|M| = k_{total}$) of top edges with the highest count. We denote this set of globally relevant features as $\mathcal{E}_{\text{context}}\subseteq \mathcal{E}$. These features provide the contextual information needed when defining the set $E$ for the symbolic learner.
%



\subsection{Symbolic Learning} 
The output features $X^{(F)}$ generated by the Feature Selector need to be represented in the symbolic form accepted by the symbolic learner. In general, any symbolic machine learning system that can induce knowledge from noisy data could be used. In all three instances of our LearnAD approach, we use FastLAS because of its scalability, speed, and robustness to noisy examples. The NKG automatically instantiates the learning task for FastLAS. Recall that FastLAS requires, as a learning task, a tuple $\langle B,S_M,E\rangle$. The $S_M$ is automatically generated by the NKG as a set of rules whose head atom is the predicate $AD$ and the body conditions are all possible combinations of atoms of the form $connection(region(i), region(j), V_{strength})$, for every $(i,j)$ edge in the selected set $M$, and comparator operators ($\geq$, $>$, $<$, $\leq$) for the variable $V_{strength}$. The set $E$ is generated by creating for each datapoint $(X, AD)$ a WCDPI example $e$ defined as $\langle e_{id}, e_{pen}, \langle \{AD\},\{CN\}\rangle, e_{ctx}\rangle$ where $e_{pen}$ is an integer greater than $0$, and 
the context $e_{ctx}$ is defined as follows:
\[
e_{ctx} = \left\{ \texttt{connection}(\texttt{region}(i), \texttt{region}(j), \texttt{strength}(i,j)) \;\middle|\; (i, j) \in \mathcal{E}_{\text{ctx}} \right\},
\]
where \((i, j)\) indexes node pairs selected according to the masked node feature matrix. Here, the node feature matrix is treated as a binary mask over nodes, and \(\mathcal{E}_{\text{ctx}} \subseteq V \times V\) denotes the set of node pairs such that both \(i\) and \(j\) are among the selected context nodes. The \texttt{strength} term refers to the edge weight between region \(i\) and region \(j\). Examples for CN patients are generated in a similar way, with the difference that their respective $e^{inc}=\{CN\}$ and $e^{exc}=\{AD\}$.
As discussed, noisy data is common in neuroimaging data, and a finite noisy penalty is applied to account for this. A uniform penalty $e_{pen}$ is applied across all examples in $E$, as the distribution of classes is equal. A non-uniform penalty could have been implemented in examples with low prediction confidence. Similarly, a penalty could have downweighted brain regions less frequently observed during the explanation stage.

\section{Experiments and Results}
\label{sec:discussion}
We use our LearnAD approach to learn interpretable rules that can differentiate between AD and CN. Patients with AD exhibit neurodegenerative changes in brain structure (e.g., reduced volume and cortical thinning). Our clinical hypothesis is that alterations in structural connectivity provide a discriminative signal for identifying AD manifestation. Our research question is whether symbolic machine learning, such as FastLAS, can learn interpretable, generalisable rules in this complex clinical setting, achieving comparable accuracy while improving interpretability relative to established statistical baselines.

We perform the task on structural connectomes derived from T1W and DWI MRI from ADNI. Each connectome comprises $84$ ROIs; self-connections (diagonals) are set to $0$. Because probabilistic tractography can introduce spurious edges, we apply proportional thresholding—retaining connections that occur in a sufficient proportion of subjects. We use a strict threshold that preserves $30\%$ of edges, yielding a network sparsity of $70\%$. We favour stringent thresholds, which have been shown to elicit stronger age associations \citep{Buchanan2020}.

We train statistical models on the full brain connectomes and evaluate them with repeated stratified cross-validation. We perform 10 repeats with distinct random seeds and report the mean training and validation scores across repeats. In each repeat, the seed is used to generate a new 90\% subsample of the dataset and to derive the fold assignments. All subsampling and fold assignments are stratified by diagnosis, sex, and MRI machine manufacturer \citep{Kan2022}. Consequently, each repeat operates on a distinct subsample and fold configuration, helping to minimise noise from manufacturer-induced artefacts. We evaluate Decision Tree (DT), Random Forest (RF), Graph Convolutional Network (GCN), and an SVM baseline (see Table~\ref{table:full_features}).

For the Feature Selector, for each DT and RF we pass the top $k_{global}$ features to the Network-to-Knowledge Generator (NKG). For the GCN, we extract $k_{instance}$ features per training graph, then retain the top $k_{total}$ edges globally by frequency across all training graphs. The NKG then constructs the set $E$ of examples for FastLAS. Each example’s context comprises facts encoding the two ROIs and the associated connectivity strength; strengths are rounded to four decimal places, then scaled by $1000$ (as the solver cannot handle real numbers) to improve scalability. Once an ASP program $H$ is learned, we use Clingo~\citep{DBLP:journals/corr/GebserKKS17} at inference time to evaluate the accuracy of $H$. To further improve scalability, we partition AD examples into disjoint subsets; each AD subset is paired with the full CN set to form separate learning tasks. We set a base noisy penalty $e_{pen}=1$ per example and rescale AD penalties by the CN-to-AD ratio in each task to balance positive and negative cases, yielding an effective uniform penalty across classes. The final hypothesis for each configuration is the union of rules learned across these tasks. For LearnAD(DT), $k_{global}=3$, and the number of disjoint AD sets is $3$. For LearnAD(RF), $k_{global}=6$, and the number of disjoint AD sets is $4$. For LearnAD(GCN), $k_{instance}=10$, $k_{total}=4$, and the number of disjoint AD sets is $3$. Performance of DT$^{*}$, and RF$^{*}$, (where $^{*}$ denotes a model applied on the extracted feature set) and their corresponding symbolic models on the selected features is reported in Table~\ref{table:subset}. 


\begin{table}[hbtp]
  \centering
  \begin{minipage}{0.5\textwidth}
    \centering
    \caption{Statistical ML ($X$)}
    \label{table:full_features}
    \begin{tabular}{lcc}
      \toprule
      \bfseries Model & \bfseries ACC (\%) \\
      \midrule
      SVM & 66.35 {\scriptsize$\pm$ 1.42} \\
      RF & 69.90 {\scriptsize$\pm$ 1.61}  \\
      DT & 58.58 {\scriptsize$\pm$ 2.66}  \\
      GCN & 68.94 {\scriptsize$\pm$ 2.00}  \\
      \bottomrule
    \end{tabular}
  \end{minipage}%
  \hfill
  \begin{minipage}{0.5\textwidth}
    \centering
    \caption{Statistical ML \& FastLAS ($X^{(F)}$)}
    \label{table:subset}
    \begin{tabular}{lcc}
      \toprule
      \bfseries Model & \bfseries ACC (\%)  \\
      \midrule
      RF$^{*}$ & 70.70 {\scriptsize$\pm$ 1.73}  \\
      LearnAD(RF) & 65.65 {\scriptsize$\pm$ 2.55}  \\
      \midrule
      DT$^{*}$ & 63.86 {\scriptsize$\pm$ 3.49}  \\
      LearnAD(DT) & 65.72 {\scriptsize$\pm$ 2.38} \\
      \midrule
      GCN$^{*}$ & --- \\
      LearnAD(GCN) & 62.97 {\scriptsize$\pm$ 2.62} \\
      \bottomrule
    \end{tabular}
  \end{minipage}
\end{table}

DTs are prone to overfitting; accordingly, our Feature Selector regularises the model via explicit feature selection, improving DT accuracy from the full network to DT$^{*}$ by $5.28\%$. RFs, as ensembles, are less susceptible to overfitting, so the gain for RF$^{*}$ is negligible. We use DT$^{*}$ and RF$^{*}$ as baselines for comparison with the symbolic models. LearnAD(DT) achieves accuracy comparable to DT$^{*}$, but DT$^{*}$ produces decision rules with bodies of up to $8$ conditions as determined by the maximum depth of the tree. In contrast, FastLAS favours compressed optimal hypotheses, yielding much shorter rules in $H$ that are more human-interpretable. RF$^{*}$ performs $5.05\%$ better than its symbolic counterpart; however, we consider this a reasonable trade-off for improved interpretability. We do not provide a GCN$^{*}$ baseline on the selected features because the resulting graphs are too sparse. LearnAD(GCN) is $5.97\%$ lower in accuracy than the GCN trained on the full network, suggesting that most of the GCN’s predictive signal can be captured by simple rules. The residual gap may reflect noise introduced by \textit{GNNExplainer} and the Feature Selector or, alternatively, the intrinsic difficulty of distilling message passing and convolutional operations into concise logical constraints. On interpretability, defined by the total number of atoms, our best-performing model has $23.48 \pm 1.6$. By comparison, the DT has $250.91 \pm 38.2$ atoms and the RF has $797.00 \pm 5.35$ atoms.

We display a representative set of rules learned by LearnAD(DT) during a learning task in Figure~\ref{fig:rules}. Across all random seeds, rules involving the left temporal pole–left hippocampus connection appear consistently, and rules involving the right precuneus–right superior parietal connection occur in approximately $80\%$ of seeds. This stability across trials suggests that these connections are reliably associated with AD status in our cohort. The semi-parametric rules learned by FastLAS encode subject-level biomarkers over connection strengths, providing relative thresholds. The observed heterogeneity of AD yields multiple threshold boundaries across subjects, potentially reflecting distinct patterns of degradation or different disease subtypes. Notably, the left hippocampus, implicated in the formation of new memories, exhibits significant atrophy in AD~\citep{Pusparani2025}, and reduced connectivity between the right precuneus and right superior parietal cortex is consistent with memory impairment observed in old-age AD patients~\citep{Prawiroharjo2020}.

\section{Related Work}
\label{sec:related}
Machine learning has been applied to AD with a range of neuroimaging modalities and clinical data. Data modalities include structural MRI \citep{Gam_Disentangled_MICCAI2024}, functional MRI \citep{Qiu_Towards_MICCAI2024}, PET data \citep{Castellano2024}, genomic data \citep{Jasodanand2025} and proteomic data \citep{PichetBinette2024}.  
Classification tasks vary between the prediction of stages of dementia \citep{Jia_AnatomyAware_MICCAI2024}, scores in cognitive tests (MMSE) \citep{vanderVeere2024}, and amyloid/tau status \citep{Jasodanand2025}. Deep learning methods implemented in these tasks include Support Vector Machines, logistic regression, k-means, CNN, reinforcement learning, transformers and GNNs; this is not an exhaustive list \citep{Cabanillas-Carbonell2025,Kan2022,Saboo2021}.

The structural connectome defines anatomical connections in the brain through white matter projections \citep{Babaeeghazvini2021}. Accelerated white matter degeneration is observed in individuals with AD \citep{Phillips2024}. We can understand the progression of clinical-pathological correlations through changes in connectivity and function of distributed neural tracts; therefore, we posit that specific white matter connections should serve as biomarkers for the underlying pathology leading to cognitive decline \citep{Matthews2013}. In this work, we utilise structural connectomes, extracted from structural MRI and DWI via tract analysis. Machine learning applied to structural connectomes has moved into graph-based methods, due to the natural network structure of the brain \citep{Cui_2023}. Existing research has focused on developing novel graph-based architectures that learn new representations of the brain data with novel structural and positional encodings \citep{Jia_AnatomyAware_MICCAI2024,yang2025do,Kan2022}. The performance of these models varies, depending on study, dataset size, and distribution. With limited work on developing benchmark datasets stemming from necessary privacy frameworks and missing biomarkers, interpretability is essential for progression in our understanding of the pathological trajectory of neurodegenerative diseases such as AD.

Interpretability of these graph-based methods has been limited to out-of-the-box explainers such as gradient-based saliency methods \citep{Wan_Selfguided_MICCAI2024}. The integration of symbolic architectures into deep graph learning is less studied. Architectures that utilise symbolic methods in the explanation of GNNs include GLGExplainer \citep{azzolin2023globalexplainabilitygnnslogic}, GraphTrail \citep{armgaan2024graphtrail}, and Monotonic GNNs \citep{cucala2022explainable}. Neuro-symbolic AI aims of marrying the reasoning capabilities of symbolic AI with the powerful learning capabilities of connectionist modern deep learning \citep{ott:hal-04136556}. Advancements in neuro-symbolic learning include the injection of neural inferences to facilitate symbolic inference \citep{Cunnington2023}, the improvement in neural training through integration of symbolic knowledge and reasoning, and the mutual interaction of separate neural and symbolic components \citep{Manhaeve2018}. Prior work with ILP in the analysis of brain networks is nonexistent as far as the authors are aware. Therefore, this work represents the first step in learning interpretable and logically consistent rules for connection strengths implicated in the development of AD.

\section{Conclusions}
This paper presents a novel neuro-symbolic approach based on a SOTA symbolic machine learning system for investigating biomarkers present in AD-affected brain networks. We have compared three statistical machine learning components, DT, RF, and GCN, as neuro-components for learning relevant search spaces. Evaluation of the symbolic models highlights the ability of symbolic machine learning to learn interpretable rules that are comparable to statistical models. We have been able to generate a hypothesis for subgraphs extracted from a GNN. This work provides a baseline for which we can extend the work with time series data of structural MRI, as well as the inclusion of more imaging modalities, such as functional MRI and PET data.
\label{sec:conclusions}

\begin{ack}
T. Andrews is supported by UK Research and Innovation [UKRI Centre for Doctoral Training in AI for Healthcare grant number EP/S023283/1].

\end{ack}

{\small
\bibliographystyle{unsrt}   
\bibliography{neurips_bib}
}

\medskip



\appendix

\section{Technical Appendix and Supplementary Materials}

\subsection{Hyperparameters and Training Details}
We evaluate Decision Tree (DT), Random Forest (RF), Graph Convolutional Network (GCN), and an SVM baseline (see Table~\ref{table:full_features}). For scikit-learn baselines (SVM, DT, RF; v1.2.1), we use defaults unless noted. The DT uses \texttt{max\_depth}=8, \texttt{criterion}='gini', and \texttt{min\_samples\_split}=2; \texttt{random\_state} varies per repeat. The RF uses \texttt{n\_estimators}=100, \texttt{max\_depth}=8, and \texttt{max\_features}='sqrt', with \texttt{random\_state} varied per repeat. The SVM uses scikit-learn defaults unless otherwise specified.

GCN experiments run on an NVIDIA GeForce RTX 3080 GPU. We stack two GCN layers with hidden dimension 32, each followed by ReLU and $\ell_2$ normalisation; input node features use dropout 0.1. Graph-level embeddings are formed by concatenating global sum and max pooling and passed to a linear classifier. Training uses 100 epochs, batch size 8, and Adam optimizer with learning rate $5 \times 10^{-3}$.

We used the \texttt{torch\_geometric.explain} implementation of \textit{GNNExplainer} provided in PyTorch Geometric (v2.5.2). The explainer was trained for 90 epochs with a learning rate of 0.1. We employed an explanation type of "phenomenon" to characterise the patterns captured by the model, and specified an edge mask type of "object" to identify the most influential edges. The node mask type was disabled (None).
\clearpage
\subsection{LearnAD(DT) Rule Visualization}
\begin{figure}[H]
  \centering
  \includegraphics[width=1\linewidth]{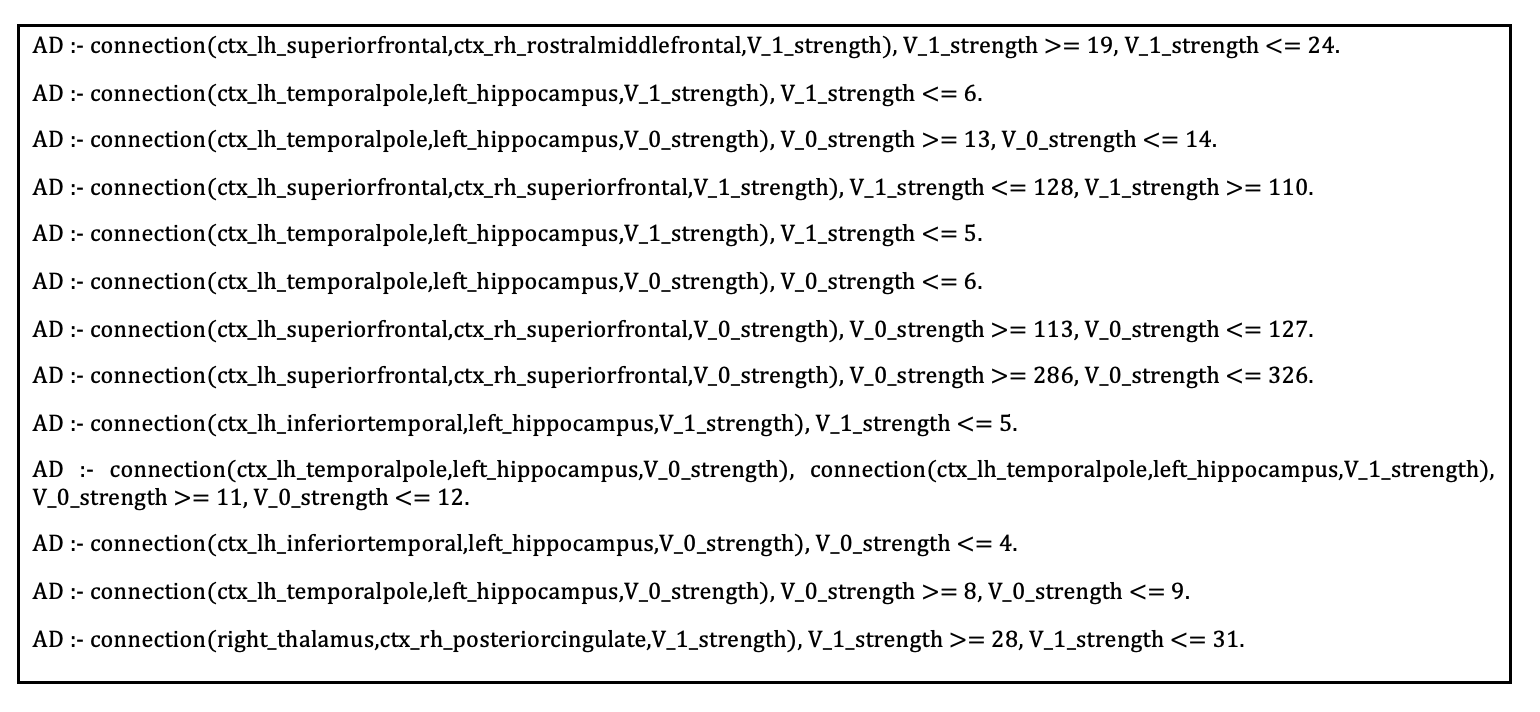}
  \caption{Representative rules learned by LearnAD(DT).}
  \label{fig:rules}
\end{figure}

We show the representative rules learned by LearnAD(DT) in Figure~\ref{fig:rules}.


\subsection{Inductive Bias and Example Construction}
\begin{figure}[H] 
  \centering
  \includegraphics[width=1\linewidth]{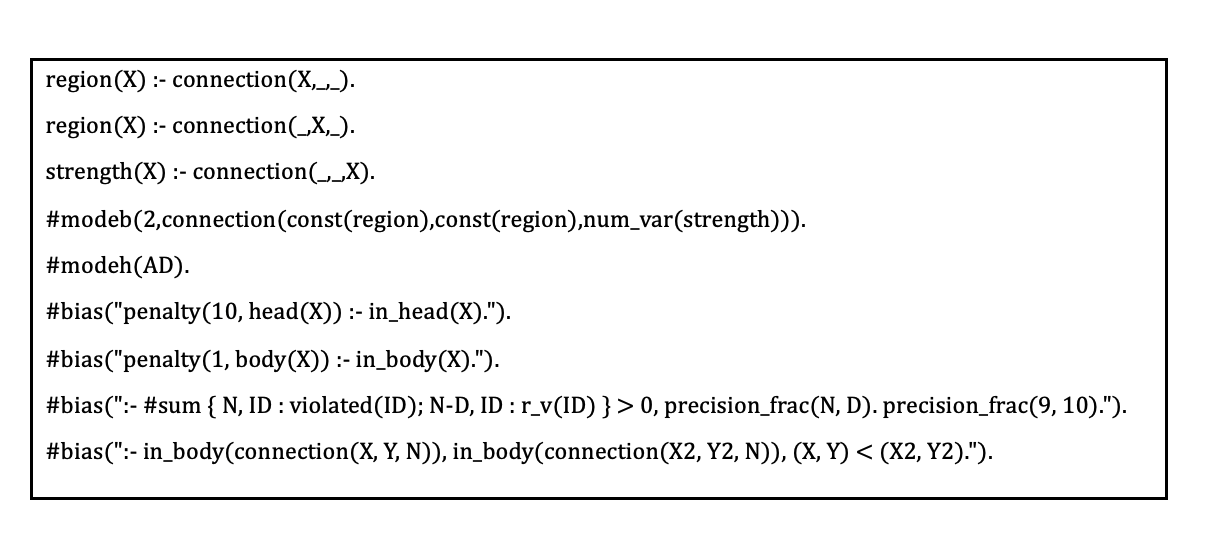}
  \caption{Inductive bias used in a FastLAS task (examples redacted for privacy).}
  \label{fig:ib}
\end{figure}

We illustrate the inductive bias used in FastLAS in Figure~\ref{fig:ib}.

\end{document}